\documentclass[11pt,a4paper]{article}
\usepackage[hyperref]{acl2021}
\usepackage{times}
\usepackage{latexsym}

\usepackage{microtype}

\usepackage{graphics}
\usepackage{graphicx}
\usepackage{multicol}

\usepackage{enumitem} 
\usepackage{booktabs} 
\usepackage{multirow}

\usepackage{amsmath}
\usepackage{bm}
\DeclareMathOperator*{\argmax}{arg\,max}

\usepackage{cleveref}
\crefformat{section}{\S#2#1#3} 
\crefformat{subsection}{\S#2#1#3}
\crefformat{subsubsection}{\S#2#1#3}

\aclfinalcopy 
\def\aclpaperid{2510} 

\title{How \textit{Good} Is NLP? A Sober Look at NLP Tasks through the \\
Lens of Social Impact}

\author{Zhijing Jin \\
  Max Planck Institute \& ETH Zürich \\
  \texttt{jinzhi@ethz.ch} \\
  \And
  Geeticka Chauhan \\
  MIT \\
  \texttt{geeticka@mit.edu} \\
  \AND
  Brian Tse \\
  Oxford \\
  \texttt{briantsemanhei@gmail.com} \\
  \And
  Mrinmaya Sachan \\
  ETH Zürich\\
  \texttt{msachan@ethz.ch} \\
  \And
  Rada Mihalcea \\
  University of Michigan \\
  \texttt{mihalcea@umich.edu} \\}

\date{}

\begin{document}
\maketitle
\begin{abstract}
Recent years have seen many breakthroughs in natural language processing (NLP), transitioning it from a mostly theoretical field to one with many real-world applications. Noting the rising number of applications of other machine learning and AI techniques with pervasive societal impact, we anticipate the rising importance of developing NLP technologies for social good. Inspired by theories in moral philosophy and global priorities research, we aim to promote a guideline for social good in the context of NLP. 
We lay the foundations via the moral philosophy definition of social good, propose a framework to evaluate the direct and indirect real-world impact of NLP tasks, and
adopt the methodology of global priorities research to identify priority causes for NLP research. Finally, we use our theoretical framework to provide some practical guidelines for future NLP research for social good.\footnote{Our data and code are available at \url{http://github.com/zhijing-jin/nlp4sg_acl2021}. In addition, we curate a list of papers and resources on NLP for social good at \url{https://github.com/zhijing-jin/NLP4SocialGood_Papers}.}
\end{abstract}

\section{Introduction}\label{sec:intro}

Advances on multiple NLP fronts have given rise to a plethora of applications that are now integrated into our daily lives. 
NLP-based intelligent agents like Amazon Echo and Google Home have entered millions of households \cite{voicebot2020amazon}.
NLP tools are now prevalent on phones, in cars, and in many daily services such as Google search and electronic health record analysis \cite{townsend2013natural}.

In the current COVID-19 context, NLP has already had important positive social impact in the face of a public health crisis. When the pandemic broke out, Allen AI collected the CORD-19 dataset \cite{wang2020cord} 
with the goal of helping public health experts efficiently sift through the myriad of COVID-19 research papers that emerged in a short time period. Subsequently, NLP services such as Amazon Kendra were deployed to help organize the research knowledge around COVID-19 \cite{bhatia2020aws}. The NLP research community worked on several problems like the question-answering and summarization system CAiRE-COVID \cite{su2020caire}, the expressive interviewing conversational system \cite{welch2020expressive} and annotation schemas to help fight COVID-19 misinformation online  \cite{alam2020fighting, hossain2020covidlies}.  



 
As NLP transits from theory into practice and into daily lives, unintended negative consequences that early theoretical researchers did not anticipate have also emerged, from the toxic language of Microsoft's Twitter bot Tay \cite{shah2016microsoft}, to the leak of privacy of Amazon Alexa \cite{chung2017alexa}.
A current highly-debated topic in NLP ethics is GPT-3 \cite{brown2020language}, whose risks and harms include encoding gender and racist biases \cite{bender2021dangers}. 

It is now evident that we must consider the negative and positive impacts of NLP as two sides of the same coin, a consequence of how NLP and more generally AI pervade our daily lives. The consideration of the negative impacts of AI has engendered the recent and popular interdisciplinary field of AI ethics, which puts forth issues such as algorithmic bias, fairness, transparency and equity with an aim to provide recommendations for ethical development of algorithms. 

Highly influential works in AI ethics include \cite{buolamwini2018gender, mitchell2019model, raji2020closing, chen2019can, blodgett2020language}. AI for social good (AI4SG) \cite{tomavsev2020ai} is a related sub-field that benefits from results of AI ethics and while keeping ethical principles as a pre-requisite, has the goal of creating positive impact and addressing society's biggest challenges. Work in this space includes \citet{wang2020cord, bhatia2020aws, killian2019learning, lampos2020tracking}.  

Active conversations about ethics and social good have expanded broadly, in the NLP community as well as the broader AI and ML communities. Starting with early discussions in works such as \cite{hovy2016social, leidner2017ethical}, the communities introduced the first workshop on ethics in NLP \cite{ws2017acl} and the AI for social good workshop  \cite{luck2018workshop}, which inspired various follow-up workshops at venues like ICML and ICLR. The NLP for Positive Impact Workshops \cite{field2020workshop,nlp4posimpact-2022-nlp} find inspiration from these early papers and workshops. In 2020, NeurIPS required all research papers to submit broader impact statements \cite{castelvecchi2020prestigious, gibney2020battle}.  
NLP conferences followed suit and introduced optional ethical and impact statements, starting with ACL in 2021 \cite{acl2021ethics}.

With the growing impact of our models in daily lives, we need comprehensive guidelines for following ethical standards to result in positive impact and prevent unnecessary societal harm. \citet{tomavsev2020ai} provide general guidelines for successful AI4SG collaborations through the lens of United Nations (UN) sustainable development goals (SDGs) \cite{un2015goals} and \citet{hovy2016social, leidner2017ethical} begin the ethics discussions in NLP. However, there is room for iteration in terms of presenting a comprehensive picture of NLP for social good, with an evaluation framework and guidelines.
At the moment, researchers eager to make a beneficial contribution need to base their research agenda on intuition and word of mouth recommendations, rather than a scientific evaluation framework.

To this end, our paper presents a modest effort to the understanding of social good, and sketches thinking guidelines and heuristics for NLP for social good. Our main goal is to answer the question:
\begin{quote}
    Given a specific researcher or team with skills $s$, and the set of NLP technologies $\bm{T}$ they can work on, what is the best technology $t\in \bm{T}$ for them to optimize the social good impact $I$?
\end{quote}

In order to answer this overall question, we take a multidisciplinary approach in our paper:
\begin{itemize}[nolistsep]
    \item \cref{sec:definition_of_good} relies on theories in moral philosophy to approach what is social good versus bad (i.e., the sign and rough magnitude of impact $I$ for a \textit{direct} act $a$);
    \item \cref{sec:nlp_tasks} relies on causal structure models as a framework to estimate $I$ for $t \in \bm{T}$, considering that $t$ can be an \textit{indirect} cause of impact;
    \item \cref{sec:priority} relies on concepts from global priorities research and economics to introduce a high-level framework to choose a technology $t$ that optimizes the social impact $I$;
    \item \cref{sec:nlp_examples} applies the above tools to analyze several example NLP directions, and provides a practical guide on how to reflect on the social impact of NLP.
\end{itemize}

We acknowledge the iterative nature of a newly emerging field in NLP for social good, requiring continuing discussions on definitions and the development of ethical frameworks and guidelines. 
Echoing the history of scientific development \cite{kuhn2012structure}, the goal of our work is not to provide a perfect, quantitative, and deterministic answer about how to maximize social good with our NLP applications. The scope of our work is to take one step closer to a comprehensive understanding, through high-level philosophies, thinking frameworks, together with heuristics and examples.


\section{What is social good?} \label{sec:definition_of_good}
Defining social good can be controversial. For example, if we define saving energy as social good, then what about people who get sick because of not turning on the air-conditioner on a cold day?
Therefore, social good is context-dependent, relevant to \textit{people}, \textit{times}, and \textit{states of nature} \cite{broome2017weighing}. 
This section is to provide a theoretical framework about the social impact $I$ for a direct act $a$. 

\subsection{Moral philosophy theories}\label{sec:philosophy}
We can observe that for some acts, it is relatively certain to judge whether the impact is positive or negative. For example, solving global hunger is in general a positive act. Such judgement is called intuitionalism \cite{sidgwick1874methods}, a school of moral philosophy.

There are many areas of social impact that cannot receive consensus by intuitions. To find analytical solutions to these debatable topics, several moral philosophies have been proposed.
We introduce below three categories of philosophical perspectives to judge moral laws \cite{kagan2018normative}, and provide the percentage of professional philosophers who support the theory \cite{bourget2014philosophers}:
\begin{enumerate}[nolistsep]
    \item Deontology: emphasizes duties or rules, endorsed by 25.9\% philosophers;
    \item Consequentialism: emphasizes consequences of acts, endorsed by 23.6\% philosophers;
    \item Virtue ethics: emphasizes virtues and moral character, endorsed by 18.2\% philosophers.
\end{enumerate}
Note that the above three schools, deontology, consequentialism, and virtue ethics, follows the standard textbook introductions for normative ethics in the analytic philosophy tradition. It is also possible for future research to consider different perspectives while defining social good.


\paragraph{A practical guide for using these philosophies.}
The three perspectives provide us dimensions to think about the impact $I$ of an act $a$, so that the final decision is (hopefully) more reliable than one single thought which is subject to biases. Such decomposition practices are often used in highly complicated analyses (e.g., business decisions), such as radar charts to rate a decision/candidate or SMART goals.

A practical guide for using moral philosophies to judge an act $a$ is to think along each of the three perspectives, collect estimations of how good the act $a$ is from the three dimensions, and merge them. For example, using NLP for healthcare to save lives can be good from all three perspectives, and thus it is an overall social good act.

When merging judgements from the above philosophical views, there can be tradeoffs, such as sacrificing one life for five lives in the 
Trolley problem \cite{thomson1976killing}, which scores high on consequentialism but low on deontology and virtue ethics. One solution by the moral uncertainty theory \cite{macaskill2014normative} is to favor acts with more balanced judgements on all criteria, and reject acts that are completely unacceptable on any criterion.

\subsection{Principles for future AI}
Many agencies from academia, government, and industries have proposed principles for future AI \cite{jobin2019global}, which can be regarded as a practical guide by \textit{deontology}. 
\citet{zeng2019linking} surveyed the principles of the governance of AI proposed by 27 agencies. The main areas are as follows (with keywords):
\begin{itemize}[nolistsep]
    \item Humanity: beneficial, well-being, human right, dignity, freedom, education, human-friendly.
    \item Privacy: personal information, data protection, explicit confirmation, control of the data, notice and consent.
    \item Security: cybersecurity, hack, confidential.
    \item Fairness: justice, bias, discrimination.
    \item Safety: validation, test, controllability.
    \item Accountability: responsibility.
    \item Transparency: explainable, predictable, intelligible.
    \item Collaboration: partnership, dialog.
    \item Share: share, equal.
    \item AGI: superintelligence.
\end{itemize}

\section{Evaluating the indirect impact of NLP}\label{sec:nlp_tasks}
Given the general moral guide to judge an act with \textit{direct} impacts, we now step towards the second stage -- understanding the downstream impact of scientific research which typically has \textit{indirect} impacts. For example, it is not easily tractable to estimate the impact of some linguistic theories. 
To sketch a solution, this section will first classify NLP tasks by the dimension of theory$\rightarrow$application, and then provide an evaluation framework for $I$ of a technology $t$ that may have indirect real-life impacts.
\subsection{Classifying tasks from upstream to downstream}\label{sec:stream}
\begin{figure}[h]
    \centering
    \includegraphics[width=\columnwidth]{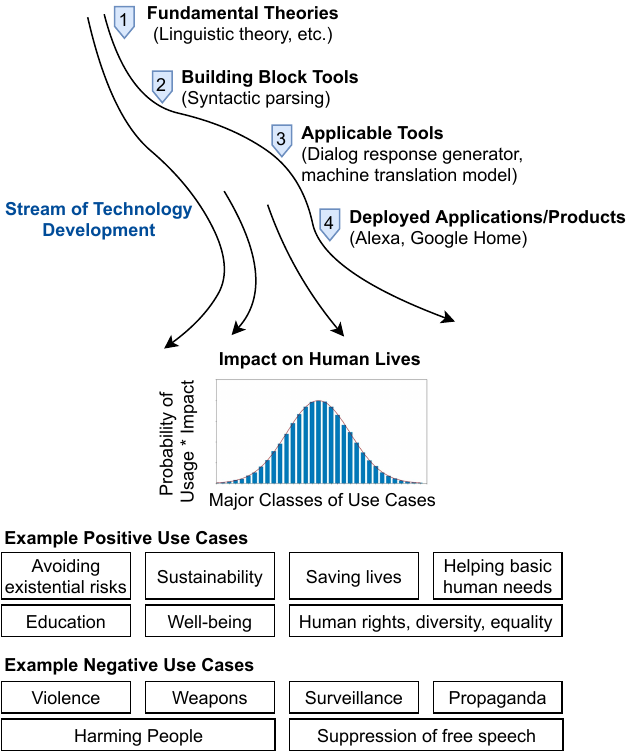}
    \caption{Stream of technology development from theory to application with end impacts. The end impacts are a distribution of use cases and their corresponding weighted impacts.}
    \label{fig:stream}
\end{figure}

To evaluate each NLP research topic, we propose four stages in the \textit{theory}$\rightarrow$\textit{application} development, as shown in Figure~\ref{fig:stream}, and categorize the 570 long papers from ACL 2020\footnote{\url{https://www.aclweb.org/anthology/events/acl-2020/\#2020-acl-main}} according to the four stages in Figure~\ref{fig:acl_stage}. Details of the annotation are in Appendix~\ref{appd:acl_annot}.
\begin{figure}
    \centering
    \includegraphics[width=0.9\columnwidth]{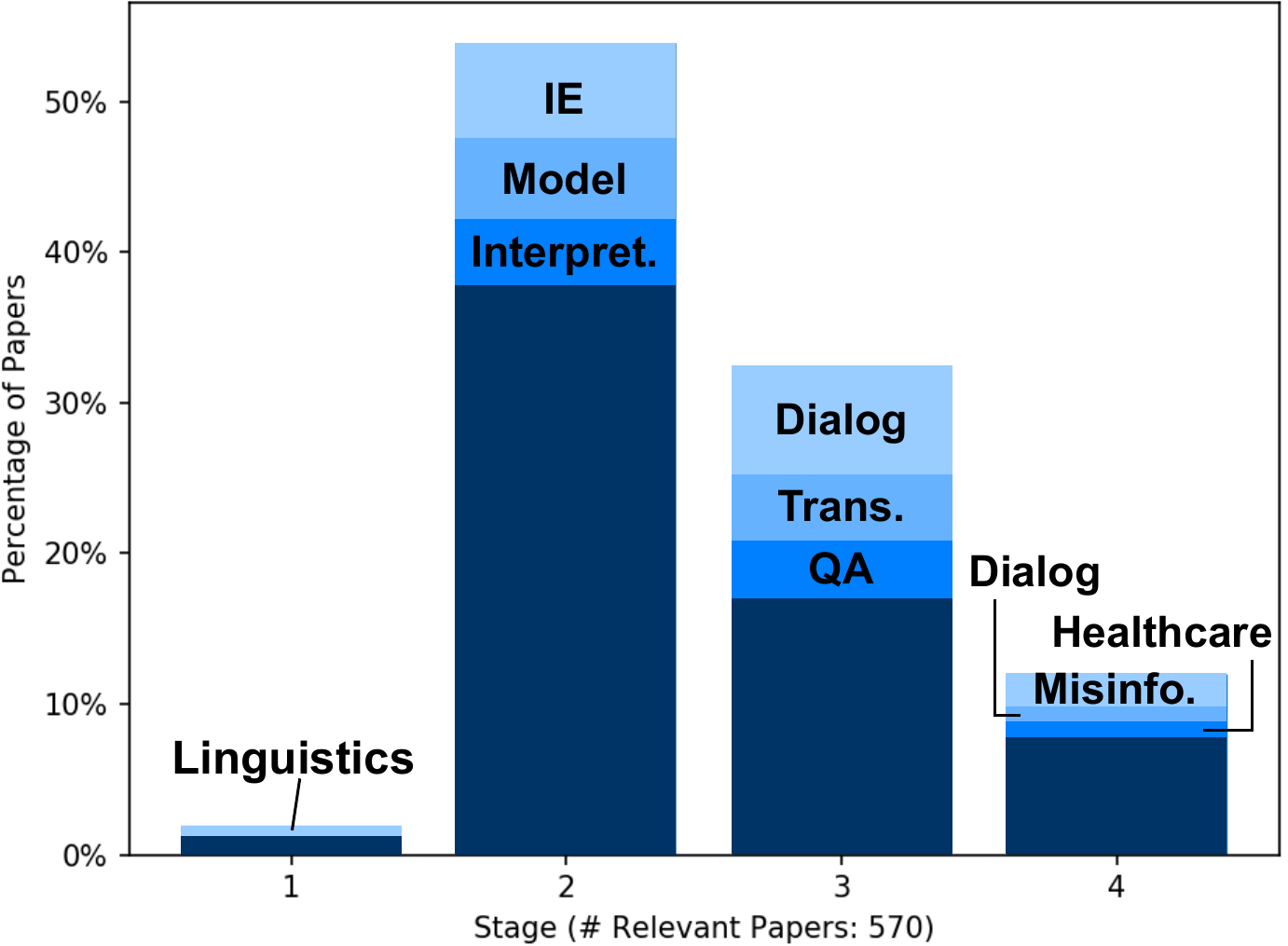}
    \caption{Distribution of ACL 2020 papers by the four stages. For each stage, we highlight the top several topics of the papers. We only list the top one topic for Stage 1 due to visual space limit. Abbreviations of technologies include Information Extraction (IE), Interpretability (Interpret.), machine translation (Trans.), question answering (QA), and misinformation (Misinfo.).}
    \label{fig:acl_stage}
\end{figure}
The four stages are as follows.

\paragraph{Stage 1. Fundamental theories.} 
Fundamental theories are the foundations of knowledge, such as
linguistic theories by Noam Chomsky.
In ACL 2020, the most prevalent topic for papers in Stage 1 is linguistics theory in Figure~\ref{fig:acl_stage}. Importantly, Stage 1's main goal is the advancement of knowledge, and to widen the potentials for later-stage research.

\paragraph{Stage 2. Building block tools.}
Moving one step from theory towards applications is the research on building block tools, which
serves as important building blocks and toolboxes for downstream technologies.  
The most frequently researched Stage-2 topics at ACL 2020 are information extraction, model design, and interpretability (in Figure~\ref{fig:acl_stage}). 

\paragraph{Stage 3. Applicable tools.}
Applicable tools are pre-commercialized NLP systems which can serve as the backbones of real-world applications. This category includes NLP tasks such as dialog response generation, question answering, and machine translation.
The most common research topics in this category are dialog, machine translation, and question answering (in Figure~\ref{fig:acl_stage}).

\paragraph{Stage 4. Deployed applications/products.}
Deployed applications often build upon tools in Stage 3, and wrap them with user interfaces, customer services, and business models.
Typical examples of Stage-4 technologies include Amazon Echo, Google Translate, and so on.
The top three topics of ACL 2020 papers in this category are ways to address misinformation (e.g., a fact checker for news bias), dialog, and NLP for healthcare.

\subsection{Estimating impact}

\paragraph{\textit{Direct} impacts of Stage-4 technologies.}
A direct impact of NLP development is allowing users more free time. This is evident in automatic machine translation, which saves the effort and time of human translators, or in NLP for healthcare, which allows doctors to more quickly sift through patient history. Automatic fake news detection frees up time for human fact-checkers, to aid them in more quickly detecting fake news through the increasing number of digital news articles being published.

The impact of more user free time is varied. In the case of healthcare, NLP can free up time for more personalized patient care, or allow free time for activities of choice, such as spending time on passion projects or more time with family. We recognize these varied impacts of NLP deployment, and recommend user productivity as one way to measure it. 

Note that there can be positive as well as negative impact associated with rising productivity, and the polarity can be decided according to Section~\ref{sec:philosophy}.
Typical positive impacts of NLP technology include better healthcare and well-being, and in some cases it indirectly helps with avoiding existential risks, sustainability, and so on. Typical negative impacts include more prevalent surveillance, propaganda, breach of privacy, and so on.
For example, intelligent bots can improve efficiency at work (to benefit economics), and bring generally better well-being for households, but they might leak user privacy \cite{chung2017alexa}.

Thus, estimating the overall end impact of a technology $t$ in the Stage 4 needs to accumulate over a set of aspects $\bm{AS}$: 
\begin{align}
    I(t) = \sum_{as\in \bm{AS}} \mathrm{scale}_{as}(t) \cdot \mathrm{impact}_{as}(t)
    ~, \label{eq:impact_stage5}
\end{align}
where $\mathrm{scale}_{as}(t)$ is the usage scale of applications of technology $t$ used in the aspect $as$, and $\mathrm{impact}_{as}(t)$ is the impact of $t$ in this aspect. 

\paragraph{\textit{Indirect} impacts of early stage technologies.}

Although the direct impact of Stage-4 technologies can be estimated by Eq.~\eqref{eq:impact_stage5}, it is difficult to calculate the impact of a technology in earlier stages (i.e., Stage 1-3).

\begin{figure}[t]
    \centering
    \includegraphics[width=0.7\columnwidth]{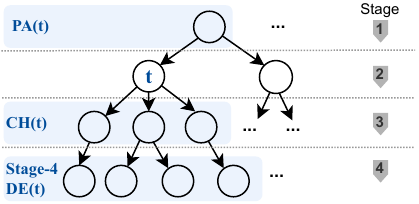}
    \caption{Viewing the theory$\rightarrow$application process with a structural causal model.}
    \label{fig:scm}
\end{figure}

We can approach the calculation of indirect impacts $I$ of an early-stage technology $t$ by a structural causal model. As shown in the causal graph $\mathcal{G}$ in Figure~\ref{fig:scm}, each technology $t$ is in a causal chain from its parent vertex set $\mathrm{PA}(t)$ (i.e., upstream technologies that directly causes the invention of $t$), to its children vertex set (i.e., downstream technologies directly resulting from $t$). Formally, we denote
a directed (causal) path in $\mathcal{G}$ as a sequence of distinct vertices $(t_1, t_2, \dots, t_n)$ such that $t_{i+1} \in \mathrm{CH}(t_i)$ for all $i=1,\dots,n-1$. We call $t_n$ a descendant of $t_1$. After enumerating all paths, we denote the set of all descendants of $t$ as $\mathrm{DE}(t)$. Specifically, we denote all descendant nodes in Stage 4 as $\mathrm{Stage}$-4 $\mathrm{DE}(t)$.

Hence, the impact of any technology $t$ is the sum of impact of all its descendants in Stage 4: 
\begin{align}
    I(t) & = \sum_{x \in \mathrm{Stage}\text{-4 }\mathrm{DE}(t)} p(x) \cdot c_x(t) \cdot I(x)
    ~,    \label{eq:impact_indirect}
\end{align}
where $p(x)$ is the probability that the descendent technology $x$ can be successfully developed, $c_x(t)$ is the contribution of $t$ to $x$, and $I(x)$ can be calculated by Eq.~\eqref{eq:impact_stage5}. This formula can also be interpreted from the light of do-calculus \cite{pearl1995causal} as $P(X|\mathrm{do}(t)) - P(X)$, for $X \in \mathrm{Stage}\text{-4 }\mathrm{DE}(t)$, which means the effect of intervention $\mathrm{do}(t)$ on Stage 4 descendants.

Note that Eq.~\eqref{eq:impact_stage5} and \eqref{eq:impact_indirect} are meta frameworks, and we leave it to future work to utilize these for assessing the social impact of their work.


\subsection{Takeaways for NLP tasks}
With the growing interest of AI and NLP publication venues (e.g., NeurIPS, ACL) in ethical and broader impact statements, it will be useful and important for researchers to have practical guidelines on  evaluating the impact of their NLP tasks.

We first introduce some thinking steps to estimate the impact of research on an NLP task $t$:
\begin{enumerate}[nolistsep,label=(S\arabic*)]
    \item Classify the NLP task $t$ into one of the four stages (\cref{sec:stream})
    \item If $t$ is in Stage 4, think of the set of aspects $\bm{AS}$ that $t$ will impact, the scale of applications, and aspect-specific impact magnitude. Finally, estimate impact using Eq.~\eqref{eq:impact_stage5}.
    \item[(S2')] If $t$ is in Stage 1-3, think of its descendant technologies, their success rate, and the contribution of $t$ to them. Finally, estimate impact using Eq.~\eqref{eq:impact_stage5} and \eqref{eq:impact_indirect}.
\end{enumerate}

Next, we introduce some high-level heuristics to facilitate fast decisions:
\begin{enumerate}[nolistsep,label=(H\arabic*)]
    \item For earlier stages (i.e., Stage 1-2), it is challenging to quantify the exact social impact. Their overall impact tends to lean towards positive as they create more knowledge that benefits future technology development. \label{item:early_stage}
    \item Developers of Stage-4 technologies should be the most careful about ethical concerns. Enumerate the use cases, and estimate the scale of each usage by thinking of the stakeholders, economic impact, and users in the market. Finally, evaluate the final impact before proceeding. (E.g., if the final impact is very negative, then abandon or do it with restrictions).
    \item For Stage-3 technologies, if their Stage-4 descendants are tractable to enumerate and estimate for their impacts, then aggregate the descendants' impacts by Eq.~\ref{eq:impact_indirect}. Otherwise, treat them like \ref{item:early_stage}.
\end{enumerate}


\section{Deciding research priority}
\label{sec:priority}

There are many directions for expansion of our efforts for social good; however, due to limited resources and availability of support for each researcher, we provide a research priority list. In this section, we are effectively trying to answer the overall question proposed in Section~\ref{sec:intro}. Specifically, we adopt the practice in the research field \textit{global priorities} (GP) \cite{macaskill2015doing,greaves2017global}. We first introduce the high-level decision-making framework in Section~\ref{sec:int}, and then formulate these principles using technical terms in Section~\ref{sec:calc_priority}.

\subsection{Important/Neglected/Tractable (INT) framework}\label{sec:int}

Our thinking framework to address the research priority follows the practice of existing cost-benefit analysis in GP \cite{macaskill2015doing,greaves2017global}, which aligns with the norms in established fields such as development economics, welfare economics, and public policy.

We draw an analogy between the existing GP research and NLP for social good. Basically, GP addresses the following problem: given, for example, 500 billion US dollars (which is the annual worldwide expenditure on social good), what priority areas should we spend on? Inspired by this practical setting, we form an analogy to NLP research efforts, namely to answer the question proposed in Section~\ref{sec:intro} about how to attribute resources and efforts on NLP research for social good.

The high-level intuitions are drawn from the Important/Neglected/Tractable (INT) framework \cite{macaskill2015doing}, a commonly adopted framework in global priorities research on social good. Assume each agent has something to contribute (e.g., money, effort, etc.). 
It is generally effective to contribute to important, neglected, and tractable areas.

\subsection{Calculation of priority}\label{sec:calc_priority}
Although the INT framework is commonly used in practice of many philanthropy organizations \cite{macaskill2015doing}, it will be more helpful to formulate it using mathematical terms and economic concepts. Note that the terms we formulate in this section can be regarded as elements in our proposed thinking framework, but they are not directly calculable.\footnote{We adapted these terms from GP. Such terms to estimate priority has been successfully used by real-world social good organizations, e.g., GiveWell, Global Priorities Institute, the Open Philanthropy Project (a foundation with over 10 billion USD investment), ReThink Priorities, 80,000 Hours Organization. In the long run, the NLP community may potentially benefit from aligning with GP's terminology. Still, we do not recommend applying our framework in high-stake settings yet, since it serves only as a starting point currently.}

Our end goal is to estimate the cost-effectiveness of contributing a unit time and effort of a certain researcher or team to research on the technology $t$. So far we have a meta framework to estimate the impacts $I$ brought by successful development of a technology $t$. And we introduce the notations in Table~\ref{tab:notation}.
\begin{table}[ht]
\small
    \centering
    \begin{tabular}{p{1cm}p{6cm}}
    \toprule
    \textbf{Notation} & \textbf{Meaning} \\ \midrule
$r$ & An NLP researcher or research group\\
$\bm{T}(r)$ & The set of NLP topics that the researcher can pursue (limited by skills, resources, and passion) \\
$t$ & An NLP technology \\
$I(t)$ & Social impacts brought by successful development of $t$ \\
$\mathrm{prog}(t)$ & The current progress of $t$\\
$p(t; r)$ & Probability that research in $t$ succeeds based on the skills of the researcher $r$\\
$p(t; r) I(t)$ & Expected social impact of the researcher $r$'s work on $t$ \\
$\Delta t(r)$ & Improvement of $t$ per unit resource (incl. time, effort, money, etc.) of the researcher $r$ \\
\bottomrule
    \end{tabular}
    \caption{Notations and their corresponding meanings used for cost-effectiveness calculation.}
    \label{tab:notation}
\end{table}

For a researcher $r$, the action set per unit resource is $\{\Delta t | t \in \bm{T}(r)\}$. Equivalently speaking, they can intervene at a node $t$ by the amount of $\Delta t(r)$ in the structured causal graph $\mathcal{G}$ in Figure~\ref{fig:scm}.

The first useful concept is $p(t; r) I(t)$, \textbf{the expected social impact} of research on a technology $t$. Here the success rate $p(t; r)$ is crucial because most research does not necessarily produce the expected outcome. However, if the impact of a technology can be extremely large (for example, prevention of extinction has impact near positive infinity), then even with a very little success rate, we should still devote considerable efforts into it.

The second concept that is worth attention is \textbf{the marginal impact} \cite{pindyck1995microeconomics} of one more unit of resources of the researcher $r$ into the technology $t$, calculated as
\begin{align}
\resizebox{0.89\hsize}{!}{%
$
\Delta I (t; r) := I(\mathrm{prog}(t)+\Delta t(r)) - I(\mathrm{prog}(t))
~.
$
}
\end{align}
For example, if the field associated with the technology is almost saturated, or if many other researchers working on this field are highly competent, then, for a certain research group, blindly devoting time to the field may have little marginal impact. However, on the other hand, if a field is important but neglected, the marginal impact of pushing it forward can be large. This also explains why researchers are passionate about creating a new research field.

The third useful concept is \textbf{the opportunity cost} \cite{palmer1999opportunity} to devote researcher $r$'s resources into the technology $t$ instead of a possibly more optimal technology $t^\star$. Formally, the opportunity cost is calculated as
\begin{align}
    t^\star (r) &:= \argmax_x \Delta I(x(r)),
    \\
    \mathrm{Cost}(t; r) &:= \Delta I(t^\star(r); r) - \Delta I (t; r)
    ~,
\end{align}
where $t^\star$ is the optimal technology that can bring the largest expected improvement of social impact.
The opportunity cost conveys the important message that we should not just do \textit{good}, but do the \textit{best}, because the difference from good to best can be a large loss.


\paragraph{Estimating the variables.}
Note that the frameworks we have proposed so far are at the meta level, useful for guiding thought experiments, and future research. Exact calculations are not possible with the current state of research in NLP for social good, although achievable in the future.

A practical insight is that NLP researchers estimate the impact of their research via qualitative explanations (natural language) or rough quantitative ones. For example, the introduction section of most NLP papers or funding proposals is a natural language-based estimation of the impact of the research. 
Such estimations can be useful to some extent \cite{hubbard2011measure}, although precise indicators of impact can motivate the work more strongly.

We can also borrow some criteria from effective altruism, a global movement that establishes a philosophical framework, and also statistical calculations of social good. One of the established metrics for calculating impact is called the ``quality-adjusted life years'' (QALYs) proposed by \citet{macaskill2015doing}. QALYs count the number of life years (calibrated by life quality such as health conditions) that an act helps to increase.

\section{Evaluating NLP tasks}\label{sec:nlp_examples}
In this section, we will first try to categorize the current state of NLP research for social good based on ACL 2020 papers, and then highlight NLP topics that are aligned with the UN's SDGs. We will conclude with a practical checklist and case studies of common NLP tasks using this checklist.
\subsection{Current state of NLP research for social good -- ACL 2020 as a case study}
\begin{figure}[t]
    \centering
    \includegraphics[width=\columnwidth]{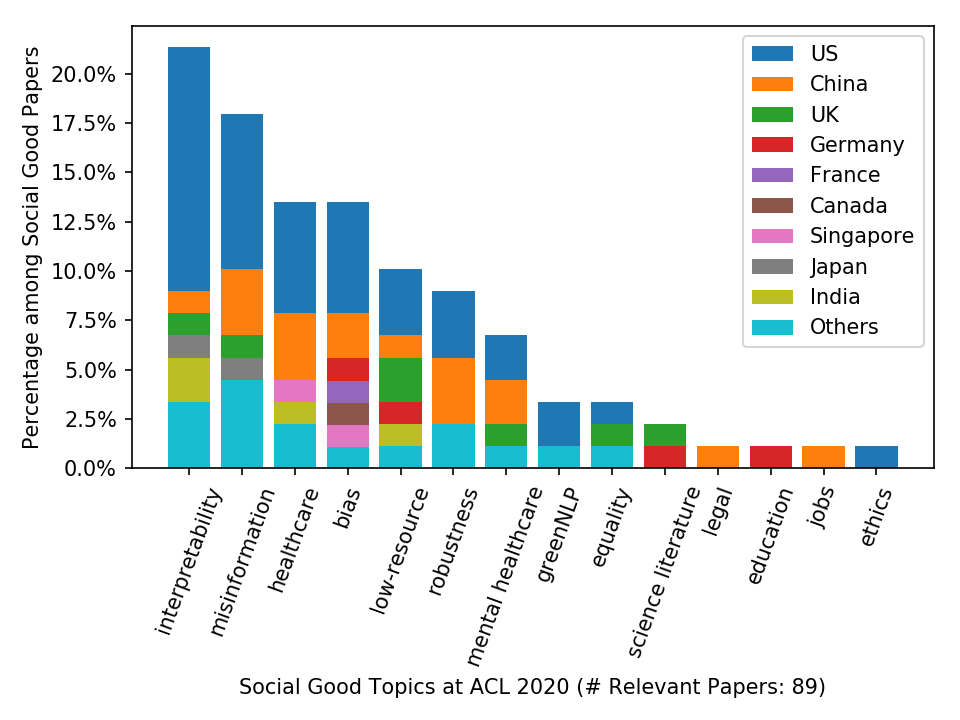}
    \caption{Social good topics at ACL 2020 by countries.}
    \label{fig:acl20_good_by_country}
\end{figure}

We want to compare the ideal priority list with the current distribution of NLP papers for social good. As a case study of the current research frontier, we plot the topic distribution of the 89 ACL 2020 papers that are related to NLP for social good in Figure~\ref{fig:acl20_good_by_country}.  We also show the portion of papers by the 10 countries with the most social-good papers. Our annotation details are in Appendix~\ref{appd:acl_annot}. 


Illustrated in Figure~\ref{fig:acl20_good_by_country}, most social-good papers work on interpretability, tackling misinformation (e.g., fact-checking for news), and healthcare (e.g., to increase the capacity of doctors). In terms of countries, the US has the most papers on interpretability, and no papers on NLP for education, NLP for legal applications, and some other topics. China has few papers on interpretability, although interpretability is the largest topic. India has no papers on fighting misinformation, although it is the second largest topic. Only 5 countries have publications across more than two social good topics.
Please refer to Appendix~\ref{appd:acl_stats} for more analyses such as social-good papers by academia vs. industries.

However, compared with the UN's SDGs \cite{un2015goals}, the current NLP research (at least in the scope of ACL conference submissions) lacks attention to other important cause areas such as tackling global hunger, extreme poverty, clean water and sanitation, and clean energy. There are also too few research papers on NLP for education, although education is the 4th most important area in SDGs.

One cause of this difference is \textit{value misalignment}. Most NLP research is supported by stakeholders and funding agencies, which have a large impact on the
current research trends or preferences in the NLP community. 
The perspective from social good with a framework to calculate the priority list has still not reached many in the NLP community. 
\begin{table}[h] 
\small
    \centering
    \begin{tabular}{ll}
    \toprule
     \textbf{Cause} & \textbf{Annual Spending (USD)} \\ \midrule
     Global R\&D & 1.5 trillion \cite{unesco2017facts} \\
     Luxury Goods & 1.3 trillion \cite{d2015luxury} \\
     US Social Welfare & 900 billion \cite{ferrara2011america} \\
     Climate Change & $>$300 billion \cite{buchner2014global} \\
     Global Poverty & $>$250 billion \cite{todd2017case} \\
     Nuclear Security & 1-10 billion \cite{todd2017case} \\
     Pandemic Prevention & 1 billion \cite{todd2017case} \\
     AI Safety Research & 10 million \cite{todd2017case}  \\
     \bottomrule
    \end{tabular}
    \caption{Annual spending of the cause areas.}
    \label{tab:misaligment}
\end{table}
\begin{table}[!thbp]
    \small
    \centering
    \begin{tabular}{p{0.13\columnwidth}p{0.8\columnwidth}}
    \toprule
    \textbf{Priority} & \textbf{Example NLP research topics} \\ \midrule
    Poverty & 
    \begin{minipage}[t]{\linewidth}
   \begin{itemize}[nosep, wide=0pt, leftmargin=*, after=\strut]
        \item Predicting poverty by geo-located Wikipedia articles \cite{sheehan2019predicting}
        \item Parsing fund applicant profiles (proposed)
    \end{itemize}
    \end{minipage}
    \\ \hline
    Hunger & \begin{minipage}[t]{\linewidth}
   \begin{itemize}[nosep, wide=0pt, leftmargin=*, after=\strut]
        \item NLP for agriculture \cite{prasad2008decision,CuiYunpeng:38}
        \item NLP for food allocation (proposed)
    \end{itemize}
    \end{minipage}
    \\ \hline
    Health \& Well-being 
 & \begin{minipage}[t]{\linewidth}
   \begin{itemize}[nosep, wide=0pt, leftmargin=*, after=\strut]
        \item NLP to analyze clinical notes \cite{dernoncourt2017neuroner, dernoncourt2017identification, luo2018segment, gopinath2020fast, leiter2020artificial,leiter2020deep}
        \item NLP for psychotherapy and counseling \cite{biester-etal-2020-quantifying,xu-etal-2020-inferring,perez-rosas-etal-2019-makes}
        \item NLP for happiness \cite{asai2018happydb,evensen2019happiness}
        \item Assistive speech generation (proposed)
    \end{itemize}
    \end{minipage}
    \\ \hline
    Education & \begin{minipage}[t]{\linewidth}
   \begin{itemize}[nosep, wide=0pt, leftmargin=*, after=\strut]
        \item NLP for educational question answering \cite{atapattu2015educational, lende2016question}
        \item Improving textbooks \cite{agrawal2010enriching}
        \item Automated grading
        \cite{madnani2018automated,taghipour2016neural}
        \item Plagiarism detection \cite{chong2010using}
        \item Tools for learners with disabilities (proposed)
    \end{itemize}
    \end{minipage}
    \\ \hline
    Equality & \begin{minipage}[t]{\linewidth}
   \begin{itemize}[nosep, wide=0pt, leftmargin=*, after=\strut]
        \item Interpretability \cite{kohn2015s, belinkov2017evaluating, 
        nie2020adversarial}
        \item Ethics of NLP \cite{hovy2016social, stanovsky2019evaluating, sap2019risk}
        \item NLP for low-resource languages \cite{
        zoph2016transfer,
        kim2017cross}
        
        \item NLP on resource-limited devices \cite{sun2020mobile}
        \item NLP tools that signal bias in human language and speech  (proposed)
    \end{itemize}
    \end{minipage}
    \\ \hline
    Clean water & \begin{minipage}[t]{\linewidth}
   \begin{itemize}[nosep, wide=0pt, leftmargin=*, after=\strut]
        \item Raising public awareness of water sanitation (proposed)
    \end{itemize}
    \end{minipage}
    \\ \hline
    Clean energy & \begin{minipage}[t]{\linewidth}
   \begin{itemize}[nosep, wide=0pt, leftmargin=*, after=\strut]
        \item Green NLP~\cite{strubell2019energy,schwartz:2020}
        \item NLP to analyze cultural values regarding climate change \cite{jiang2017comparing,keonecke2019learning}
        \item Cross-cultural models of climate change perceptions (proposed)
    \end{itemize}
    \end{minipage}
    \\
    \bottomrule
    \end{tabular}
    \caption{Top priorities and some NLP research related to each of them. This list may not be exhaustive. We also propose a high-impact research problem in each of the areas which has received less attention so far.}
    \label{tab:un_goal_and_research}
\end{table}

Although we do not have data on expenditure in each NLP subarea, we can get a glimpse of the value misalignment in general. Table~\ref{tab:misaligment} shows the annual spending of some cause areas. Note that the ranking of the expenditure does not align with our priority list for social good. For example, luxury goods are not as important as global poverty, but luxury goods cost 1.3 trillion USD each year, almost five times the expenditure in global poverty.


\subsection{Aligning NLP with social good}
In this subsection, we list the top priorities according to UN's SDGs \cite{un2015goals}. For each goal, in Table~\ref{tab:un_goal_and_research} we include examples of existing NLP research, and suggest potential NLP tasks that can be developed  (labeled as (proposed)).

    




\subsection{Checklist}
As a practical guide, we compile the takeaways of this paper into a list of heuristics that might be helpful for future practioners of NLP for social good. 
To inspect the social goodness of an NLP research direction (especially in Stage 3-4), the potential list of questions to answer is as follows:
\begin{enumerate}[nolistsep,label=(Q\arabic*)]
    \item What kind of people/process will benefit from or be harmed by the technology?
    \label{item:check_population}
    \item Does it reinforce the traditional structure of beneficiaries?
    I.e., what groups of underprivileged people can be benefited?
    (e.g., by gender, demographics, socio-economic status, country, native languages, disability type)
    \label{item:equal}
    \item Does it contribute to SDG priority goals such as poverty, hunger, health, education, equality, clean water, and clean energy? \label{item:un_goals}
    \item Can it directly improve quality of lives? E.g., how many QALYs might it result in? \label{item:qaly}
    \item Does it count as (a) mitigating problems brought by NLP, or (b) proactively helping out-of-NLP social problems? \label{item:patch_or_proactive}
\end{enumerate}


\subsection{Case studies by the checklist}\label{sec:nlp_example_impact}

We conduct some case studies of NLP technologies using the checklist.
\paragraph{Low-resource NLP \& machine translation.} 
This category includes NLP on low-resource languages, such as NLP for Filipino \cite{sagum2019ficobu,cruz2020investigating}, and MT for Haitian Creole after the 2010 Haiti earthquake \cite{lewis2010haitian}, and machine translation in general. Because this direction expands the users of NLP technologies from English-speaking people to other languages, it benefits people speaking these languages \ref{item:check_population}, and helps to narrow the gap between English-speaking and non-English speaking end users \ref{item:equal}, although it is still likely that people who can afford intelligent devices will benefit more than those who cannot. This category can contribute directly to goals such as equality and education, and indirectly to other goals because translation of documents in general helps the sharing of information and knowledge \ref{item:un_goals}. It directly improves quality of lives, for example, for immigrants who may have difficulties with the local language \ref{item:qaly}. Thus, it counts as social good category (b) in \ref{item:patch_or_proactive}.

\paragraph{Transparency, interpretability, algorithmic fairness and bias.}
Research in this direction can impact users who need more reliable decision-making NLP, such as the selection process for loans, jobs, criminal judgements, and medical treatments \ref{item:check_population}. It can shorten the waiting time of candidates and still make fair decisions regardless of spurious correlations \ref{item:equal} \ref{item:qaly}. It reduces inequality raised by AI, but not increasing equality over man-made decisions, at least by the current technology \ref{item:equal}. Thus, it is social good category (a) in \ref{item:patch_or_proactive}.


\paragraph{Green NLP.}
Green NLP reduces the energy consumption of large-scale NLP models. Although it works towards the goal of affordable and clean energy \ref{item:un_goals} by neutralizing the negative impact of training NLP models, but it does not impact out-of-NLP energy problems. Green NLP belongs to social good category (a) in \ref{item:patch_or_proactive}. It does not have large impacts directly targeted at \ref{item:check_population}, \ref{item:equal} and \ref{item:qaly}.


\paragraph{QA \& dialog.}
People who can afford devices embedded with intelligent agents can use it, which is about 48.46\% of the global population \cite{bankmycell2021smartphones} \ref{item:check_population}. So this benefits people with higher socio-economic status, and benefits English speaking people more than others, not to mention job replacements for labor-intensive service positions \ref{item:equal}. It does not contribute to priority goals except for education and healthcare for people who can afford intelligent devices \ref{item:un_goals}. Nonetheless, it can improve the quality of lives for its user group \ref{item:qaly}. It can be regarded as social good of category (b) in \ref{item:patch_or_proactive}.

\paragraph{Information extraction, NLP-powered search engine \& summarization.}
This direction speeds up the information compilation process, which can increase the productivity in many areas. About 50\% of the world population have access to the Internet and thus can use it \cite{meeker2019internet} \ref{item:check_population} \ref{item:equal}. This category indirectly helps education, and the information compilation process of other goals  \ref{item:un_goals}.
It can largely improve the lives of its user group because people gather information very frequently (e.g., do at least one Google search every day) \ref{item:qaly}. Thus, it belongs to social good category (b) in \ref{item:patch_or_proactive}.

\paragraph{NLP for social media.}
Research on social media provides tools for multiple parties. Social scientists can mine interesting trends and cultural phenomena; politicians can survey constituents' opinions and influence them; companies can investigate user interests and expand their markets \ref{item:check_population}. The caveat of dual use is large, and heavily rely on the stakeholders' intent: exploitation of the tools will lead to bleach of user privacy, and information manipulation, whereas good use of the tools can help evidence-based policy makers (social good category (a) in \ref{item:patch_or_proactive}), and help to understand the driving principles of democratic behavior and combat the mechanisms that undermine it (social good category (b) in \ref{item:patch_or_proactive}). Such diverse possibilities of parties who use them leave \ref{item:equal} and \ref{item:qaly} unanswerable. Also, this research direction has limited (and often indirect) contribution to priorities such as poverty and hunger, unless the related policies are in heat discussion online
\ref{item:un_goals}.


\section{Conclusion}
This paper presented a meta framework to evaluate NLP tasks in the light of social good, and proposed a practical guide for practitioners in NLP. We call for more attention towards awareness and categorization of social impact of NLP research, and we envision future NLP research taking on an important social role and contributing to multiple priority areas. We also acknowledge the iterative nature of this emerging field, requiring continuing discussions, improvements to our thinking framework and different ways to implement it in practice. We highlight that the goal of our work is to take one step closer to a comprehensive understanding of social good rather than introducing a deterministic answer about how to maximize social good with NLP applications.

\section*{Acknowledgments}

We thank Bernhard Schoelkopf, Kevin Jin, and Qipeng Guo for insightful discussions on the main ideas and methodology of the paper. We thank Osmond Wang for checking the economic concepts in the paper. We also thank Chris Brockett for checking many details in the paper. We thank the labmates in the LIT lab at University of Michigan, especially Laura Biester, Ian Stewart, Ashkan Kazemi, and Andrew Lee for constructive feedback. We also thank labmates at the MIT MEDG group, especially William Boag and Peter Szolovits for their constructive feedback. We also received many feedbacks based on the first version of the paper, -- we thank Niklas Stoehr for constructive suggestions to help some arguments be more comprehensive in the current version. We thank Jingwei Ni for the help with the annotation of the country and affiliation of the ACL 2020 papers. 

\section*{Ethical and societal implications}
Our paper establishes a framework to better understand the definition of social good in the context of NLP research, and lays out a recommended direction on how to achieve it. The contributions of our paper could benefit a focused, organized and accountable development of NLP for social good.
The data used in our work is public, and without privacy concerns.

\bibliographystyle{acl_natbib}
\bibliography{anthology,acl2021}

\clearpage

\appendix

\section{ACL 2020 paper annotations}\label{appd:acl_annot}

For the case study on ACL 2020 papers, such as Figure~\ref{fig:acl_stage} and~\ref{fig:acl20_good_by_country}, we collect the 570 long papers at ACL 2020. An NLP researcher with four years of research experience conducted the entire annotation, so that the categorization is consistent across all papers.\footnote{The annotation file has been uploaded to the softconf system.} 

The first annotation task is to categorize all papers into one of the four stages in the theory$\rightarrow$application development. We showed the annotator the description of the four stages in Section~\ref{sec:stream}. Next, provided with the title, abstract, and PDF of each paper, the annotator was asked to annotate which of the four stages each paper belongs to. The annotator had passed a test batch before starting the large-scale annotation.

The second annotation task is to annotate the research topics of the papers related to social good at ACL 2020. If the paper has a clear social good impact (89 out of 570 papers), the annotator needs to classify the topic of the paper into one of the given categories: bias mitigation, education, equality, fighting misinformation, green NLP, healthcare, interpretability, legal applications, low-resource language, mental healthcare, robustness, science literature parsing, and others. For the other meta information such as countries, or academia vs. industry, we decide based on the information of the leading first author.

\section{More statistics about ACL 2020 papers}\label{appd:acl_stats}

For the case study on ACL 2020 papers, we further investigate the following statistics.

\begin{figure}[h]
    \centering
    \includegraphics[width=\columnwidth]{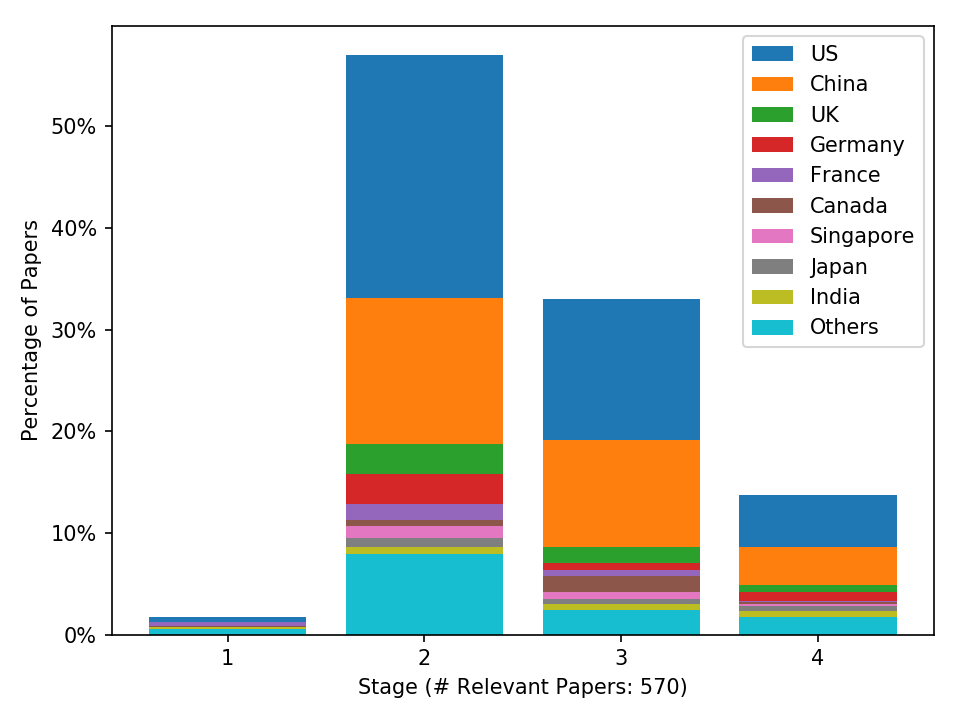}
    \caption{Stage 1-4 of ACL 2020 papers by countries.}
    \label{fig:acl20_stage_by_country}
\end{figure}

\paragraph{Stage 1-4 by countries.}
Recall that in Figure~\ref{fig:acl_stage} of the main paper, we plot the distributions of papers by the four stages, and highlight the most frequent topics in each stage. Additionally, it is also interesting to explore the distribution of stages for different countries. In Figure~\ref{fig:acl20_stage_by_country}, we have the following observations:

China does not have Stage-1 papers (i.e., fundamental theories), although it has the second largest total number of papers. The reason might be that there are not many Chinese researchers on linguistic theories who publish at English conferences.

Most countries' number of papers in the four stages follows the overall trend (i.e., Stage-2 papers $>$ Stage-3 papers $>$ Stage-4 papers $>$ Stage-1 papers), with a few exceptions. For example, China has almost the same number of papers in Stage 2 and 3, Germany has more papers in Stage 4 (i.e., deployed applications) than in Stage 3, and Canada has the most papers in Stage 3.

\paragraph{Social good topics by academia vs. industry.}
As we call for more research attention to  NLP for social good, it is important to understand the affiliations behind the current social good papers. A coarse way is to look at the affiliation of the first author, and inspect whether the main work of the paper is done by people from academia or industry.
\begin{figure}[h]
    \centering
    \includegraphics[width=\columnwidth]{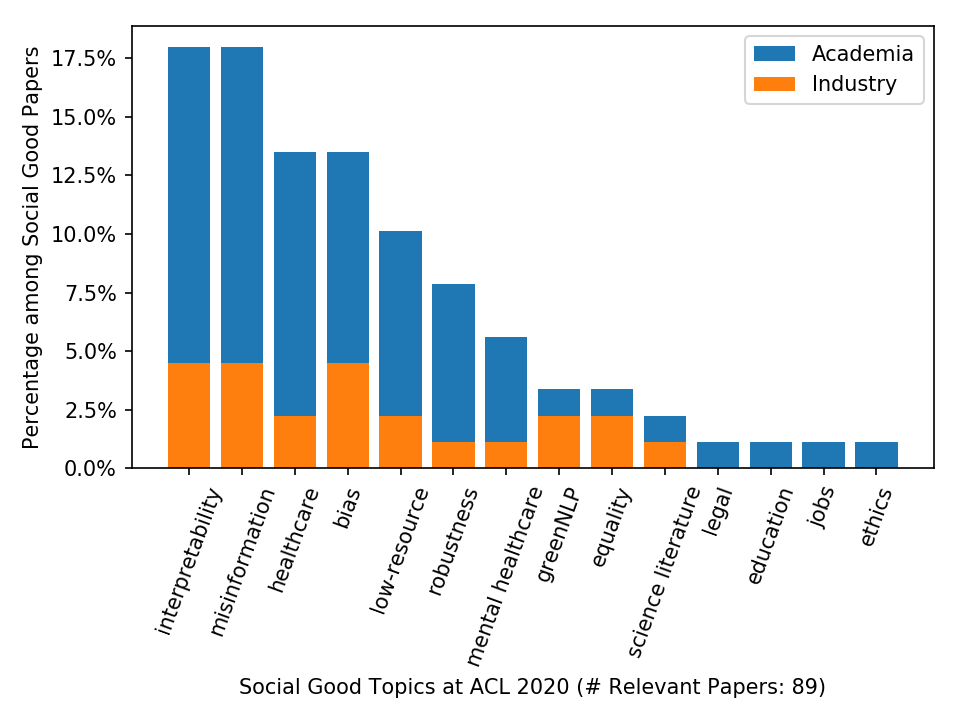}
    \caption{Social good topics at ACL 2020 by affiliations.}
    \label{fig:acl20_good_by_affiliation}
\end{figure}

As in Figure~\ref{fig:acl20_good_by_affiliation}, overall academia publishes several times more papers on social good than the industry. This ratio is higher than the average ratio of papers from academia out of all ACL 2020 papers (389 from academia out of 570). Industry does not have ACL 2020 papers on topics such as NLP ethics. Note that using statistics from ACL papers alone could be limiting because researchers in academia typically present almost all research achievements through publications, but many industry researchers do not publish in public venues such as ACL, although their research may impact various products.

\end{document}